\pdfoutput=1

\documentclass[11pt]{article}
\usepackage{acl}
\usepackage{times}
\usepackage{latexsym}
\usepackage[T1]{fontenc}
\usepackage[utf8]{inputenc}
\usepackage{microtype}
\usepackage{inconsolata}
\usepackage{graphicx}

\usepackage{multirow}
\usepackage{hyperref}
\usepackage{colortbl}
\usepackage{cite}
\usepackage{url}
\usepackage{CJK}
\usepackage{booktabs}
\usepackage{multirow}
\usepackage{xcolor}
\usepackage{titletoc}   % For appendix 
\definecolor{mygray}{gray}{.9}
\definecolor{lightblue}{RGB}{198, 226, 255}
\definecolor{lightred}{RGB}{253, 230, 224}
\definecolor{lightgreen}{RGB}{204, 232, 207}
\definecolor{lightgray}{gray}{0.95}
\definecolor{mediumgray}{gray}{0.85}
\definecolor{darkgray}{gray}{0.7}

\usepackage[most]{tcolorbox}
\usepackage{float}
\usepackage{xspace}
\usepackage[most]{tcolorbox}
\usepackage{float}
\usepackage{xspace}
\tcbset{
  aibox/.style={
    width=400pt,
    top=8pt,
    colback=white,
    colframe=black,
    colbacktitle=black, % This sets the background color of the title box
    enhanced,
    center,
    attach boxed title to top left={yshift=-0.12in,xshift=0.15in},
    boxed title style={
      boxrule=0pt,
      colframe=white, % This sets the frame color of the title box (not the text)
      fontupper=\footnotesize, % Or \small, \scriptsize, etc.
    },
  }
}
\newtcolorbox{AIbox}[2][]{aibox,title=#2,#1}

\usepackage{amsmath}
\usepackage{amssymb}
\usepackage{pifont}% http://ctan.org/pkg/pifont
\newcommand{\cmark}{\ding{51}}%
\newcommand{\xmark}{\ding{55}}%

\usepackage{mathtools}
\usepackage{amsthm}
\usepackage{wrapfig}
\usepackage{threeparttable}
\usepackage{tabularx}

\newcommand{\revise}[1]{\textcolor{black}{#1}}

\title{LongLLaVA: Scaling Multi-modal LLMs to 1000 Images Efficiently via a Hybrid Architecture}

\author{Xidong Wang$^\dagger$$^{1}$, Dingjie Song$^\dagger$$^{1,2}$, Shunian Chen$^{1}$, Junyin Chen$^{1}$, Zhenyang Cai$^{1}$, \\ \textbf{Chen Zhang}$^{3}$, \textbf{Lichao Sun}$^2$, \textbf{Benyou Wang} $^1$\thanks{Benyou is the corresponding author (\textit{wangbenyou@cuhk.edu.cn}); $^\dagger$ means contributing equally.} \\
 $^1$ The Chinese University of Hong Kong, Shenzhen\\
 $^2$ Lehigh University $^3$ Meituan\\
\url{https://github.com/FreedomIntelligence/LongLLaVA}
}

\begin{document}

\maketitle

\begin{abstract}

Expanding the long-context capabilities of Multi-modal Large Language Models~(MLLMs) is critical for advancing video understanding and high-resolution image analysis. Achieving this requires systematic improvements in model architecture, data construction, and training strategies, particularly to address challenges such as performance degradation with increasing image counts and high computational costs. In this paper, we propose a hybrid architecture that integrates Mamba and Transformer blocks, introduce data construction methods that capture both temporal and spatial dependencies, and employ a progressive training strategy. Our released model, LongLLaVA (\textbf{Long}-Context \textbf{L}arge \textbf{L}anguage \textbf{a}nd \textbf{V}ision \textbf{A}ssistant), demonstrates an effective balance between efficiency and performance. LongLLaVA achieves competitive results across various benchmarks while maintaining high throughput and low memory consumption. Notably, it can process nearly one thousand images on a single A100 80GB GPU, underscoring its potential for a wide range of multi-modal applications.
\end{abstract}

\section{Introduction}

% Background
The rapid advancement of MLLMs~\citep{liu2024llavanext, liu2023improvedllava, internlmxcomposer2, chen2024allava} has demonstrated their remarkable capabilities across various applications~\citep{chu2024mobilevlm, yang2023appagent, wu2023next, chen2024huatuogpt}. However, multi-image scenario remain an important yet to-be-explored aspect. In particular, expanding the context of MLLMs to understand longer videos~\citep{damonlpsg2023videollama, damonlpsg2024videollama2}, higher-resolution images~\citep{xu2024llavauhdlmmperceivingaspect, wu2023vguidedvisualsearch}, and make decisions based on more historical messages~\citep{wang2024mobile2,liu2024visualagentbenchlargemultimodalmodels} is crucial for enhancing user experience~\citep{li2024multimodal} and further broadening MLLMs' application scope~\citep{apple-intelligence-foundation-language-models}. 

% why this is challenging
However, extending the context length of MLLMs to improve their usability poses challenges related to degraded performance and high computational costs when processing more images. 
% Recent research based on the LLaVA architecture~\citep{liu2023llava} has primarily focused on the former. 
To maintain the performance in longer context, some studies~\citep{zhang2024internlmxcomposer25versatilelargevision, zhao2024omchatrecipetrainMultimodal} have concentrated on curating long-context training data involving multiple images to enhance performance. Additionally, other research efforts have explored new training strategies~\citep{liu2024worldmodelmillionlengthvideo,zhang2024longcontexttransferlanguage, li2024llava, zhang2024llavanextvideo} to mitigate performance declines.  
Regarding the issue of high computational costs, \citet{xue2024longvilascalinglongcontextvisual} have made strides in improving multi-node efficiency by reducing communication costs. However, a significant gap persists in accelerating core on-node computation for long visual contexts without sacrificing performance. An integrated architectural solution addressing both performance and efficiency is thus needed.
% To enable MLLMs to effectively manage multi-modal long-context scenarios while serving a larger community at a relatively low cost, a systematic solution that balances performance and computational cost is essential.

\begin{table*}[t]
\centering
\footnotesize
% \vspace{-2mm}
\addtolength\tabcolsep{-3.5pt} % Kept from the first table
\centering
% Corrected column specifier: l l c (original) | c (new ICL) c c c c (original Few-shot) | l (new Compute Complexity) c c c c (original Efficiency)
\begin{tabular}{llc|c cccc|l cccc}
\toprule
% Headers swapped: ICL is now 4th, Compute Complexity is now 6th (conceptually)
\multirow{3}{*}{\textbf{Arch.}} & \multirow{3}{*}{\textbf{Model}}& \multirow{3}{*}{\shortstack{\textbf{Active}\\\textbf{Param.}}} & \multirow{3}{*}{\textbf{\textit{ICL}}} & \multicolumn{4}{c|}{\textbf{\#Few-shot of VL-ICL}} & \multirow{3}{*}{\shortstack{\textbf{\textit{Compute}}\\\textbf{\textit{Complexity}}}} & \multicolumn{4}{c}{\textbf{100K Token (Efficiency)}}\\
& & & & 1 & 2 & 4 & 5 & & \shortstack{Prefill\\(s)} & \shortstack{TP\\(tokens/s)} & \shortstack{Mem.\\(GB)} & \shortstack{\revise{Max TP}\\(tokens/s)}\\
\midrule
% Cobra  & Mamba & 3B & 48.7 & 50.3 & 51.0 & 51.5 & 10.2 & 42.7 & 29.9 & 192.1\\ % This line was already commented
% Data swapped for ICL and Compute Complexity columns
Mamba&  \revise{Falcon-mamba-V} & 7B& \xmark & 49.0& 51.9& 52.4& 53.2& \textbf{\textit{Linear}} & 14.3 & 72.6 & 32.1 & 170.3 \\
Transformer & LLaVA-1.5 & 13B & \cmark & 50.0 & 52.3 & 54.6 & 58.9 & \textit{Quadratic} & 34.0 & 14.7 & 79.4  & 14.7\\ \midrule
\rowcolor{gray!15} Hybrid & LongLLaVA-9B & 9B & \cmark & 51.6 & 57.8 & 58.4 & 60.2 & \textbf{\textit{Quasi-Linear}} & 16.5 & 62.1 & 38.7  & 155.2\\
\rowcolor{gray!15} Hybrid & LongLLaVA-A13B & 13B & \cmark &52.3 & 59.0 & 59.0 & 61.3 & \textbf{\textit{Quasi-Linear}} & 25.5 & 37.6 & 79.1  & 37.6\\
\bottomrule
\end{tabular}
% \vspace{-2mm}
\caption{Model Architectures Analysis: ICL Capability, and Efficiency. ICL performance is reported using VL-ICL~\citep{zong2024vliclbenchdevildetails} with varying numbers of examples. Efficiency metrics for processing 100K tokens include Prefill time (Prefill), Throughput (TP), Memory usage (Mem.). The Mamba architecture is represented by Falcon-mamba~\citep{zuo2024falcon}, the largest publicly available pure Mamba LLM. Details are in Appendix~\ref{sec:hybrid}.}
% \vspace{-2mm}
\label{tab:arch2}
\end{table*}

To tackle these challenges, we propose \textbf{LongLLaVA}, featuring a hybrid architecture for efficient acceleration. Our solution focuses on three aspects: \textit{Multi-modal Architecture}, \textit{Data Construction}, and \textit{Training Strategy}.

\begin{itemize}
\item \textbf{Multi-modal Architecture}: We use a hybrid Transformer-Mamba design and 2D pooling to compress image tokens, reducing computation while maintaining performance.
\item \textbf{Data Construction}: We create task-specific formats to help the model distinguish temporal and spatial relationships between images.
\item \textbf{Training Strategy}: We implement a three-stage adaptation process to enhance model’s multi-modal long-context capabilities.
\end{itemize}

Experiemntal results show that LongLLaVA excels in understanding multi-modal long contexts with high efficiency. It leads in retrieval, counting, and ordering tasks in VNBench~\citep{zhao2024needlevideohaystackscalable} and achieves nearly 100\% accuracy with 1,000 images on a single 80GB GPU for Needle-In-A-Haystack evaluation~\citep{zhang2024longcontexttransferlanguage}.
% Contributions
% Our summarized contributions are as follows:
% \begin{itemize}
%     \item We introduce LongLLaVA, a solution optimized through data construction, training strategies, and multi-modal architecture, effectively balancing performance and efficiency. To the best of our knowledge, this is the first hybrid architecture for MLLMs.
%     \item LongLLaVA demonstrates exceptional performance in multi-modal long-context understanding, excelling in retrieval, counting, and ordering tasks.
%     In our commitment to transparency and community research, we will open source all models, codes, and datasets associated with LongLLaVA.
% \end{itemize}

\section{Background}

\subsection{The Computational Bottleneck in Multi-Image Architectures}

While open-source Multimodal Large Language Models (MLLMs) have demonstrated impressive capabilities on single-image tasks, often matching their closed-source counterparts~\citep{bai2023qwen,li2024llava,zhang2024internlmxcomposer25versatilelargevision,openai2024gpt4technicalreport,geminiteam2024gemini15unlockingmultimodal}, a significant performance disparity arises in multi-image scenarios~\citep{song2024milebenchbenchmarkingmllmslong}. This challenge stems from a fundamental computational bottleneck: the processing of excessively long visual token sequences. Standard visual encoders, such as CLIP~\citep{radford2021learningtransferablevisualmodels}, transform each image into a large set of tokens. As the number of input images scales, the length of this token sequence increases linearly, quickly overwhelming the fixed context window and computational budget of the language model. For instance, representing a mere three-minute video at one frame per second can generate a sequence exceeding 100,000 tokens, imposing prohibitive demands on both memory and processing power.

To address this scalability issue, prevailing strategies rely on visual compression~\citep{chen2023minigpt,zhang2024longcontexttransferlanguage,xu2024pllava}. These techniques mitigate the computational load by reducing the token count before it is passed to the LLM. However, this approach introduces a critical trade-off, as compression is inherently lossy. It sacrifices the fine-grained, high-frequency details within each image that are essential for nuanced understanding. This forces existing architectures into a difficult dilemma of choosing between the unsustainable computational expense of full-fidelity processing and a performance ceiling imposed by irreversible information loss. This architectural impasse, the inability to achieve both scalability and high fidelity simultaneously, serves as the primary motivation for our work and compels the exploration of a new paradigm.

\subsection{Motivation of Hybrid Architecture}
\paragraph{Architectural Strengths and Limitations} As shown in Table~\ref{tab:arch2}, Transformer architectures face significant computational challenges due to the quadratic complexity with sequence length. This inefficiency becomes a bottleneck in long-context scenarios, requiring high memory and computation resources. Mamba architectures address this issue with their linear computational complexity, making them significantly more efficient. However, they exhibit weaknesses in In-Context Learning~(ICL) tasks, particularly those involving complex retrieval or reasoning~\citep{park2024mambalearnlearncomparative}. These limitations may attributed to Mamba’s reliance on reduced attention mechanisms~\citep{olsson2022incontextlearninginductionheads}, which constrain its ability to learn contextual patterns effectively. Although explicit training allows the Mamba model to execute basic ICL tasks, it falls short of leveraging the full potential of the parameter capacity and the available training data~\citep{dao2024transformersssmsgeneralizedmodels}.

\paragraph{Synergistic Advantages of Hybrid Architecture} Recent advancements have demonstrated the potential of hybrid Mamba-Transformer architectures, which integrate Mamba’s efficiency with the robust ICL capabilities of Transformers~\citep{dao2024transformersssmsgeneralizedmodels, wang2024mamballamadistillingaccelerating}. Comparative experiments show that these hybrids achieve superior performance on ICL tasks and maintain computational efficiency. For instance, Jamba~\citep{lieber2024jambahybridtransformermambalanguage}, a hybrid model, can process 256K tokens with only 4GB of KV-Cache memory, far surpassing the capabilities of Mixtral~\citep{jiang2024mixtralexperts}, which has the same activation parameters. As shown in Table~\ref{tab:arch2}, this balance between effectiveness and efficiency makes hybrid architectures an ideal solution for long-context multimodal tasks, addressing both computational and functional limitations. Experimental details are in Appendix~\ref{sec:hybrid}.

\begin{figure*}[t]
\centering
% \vspace{-2mm}
\includegraphics[width=0.85\linewidth]{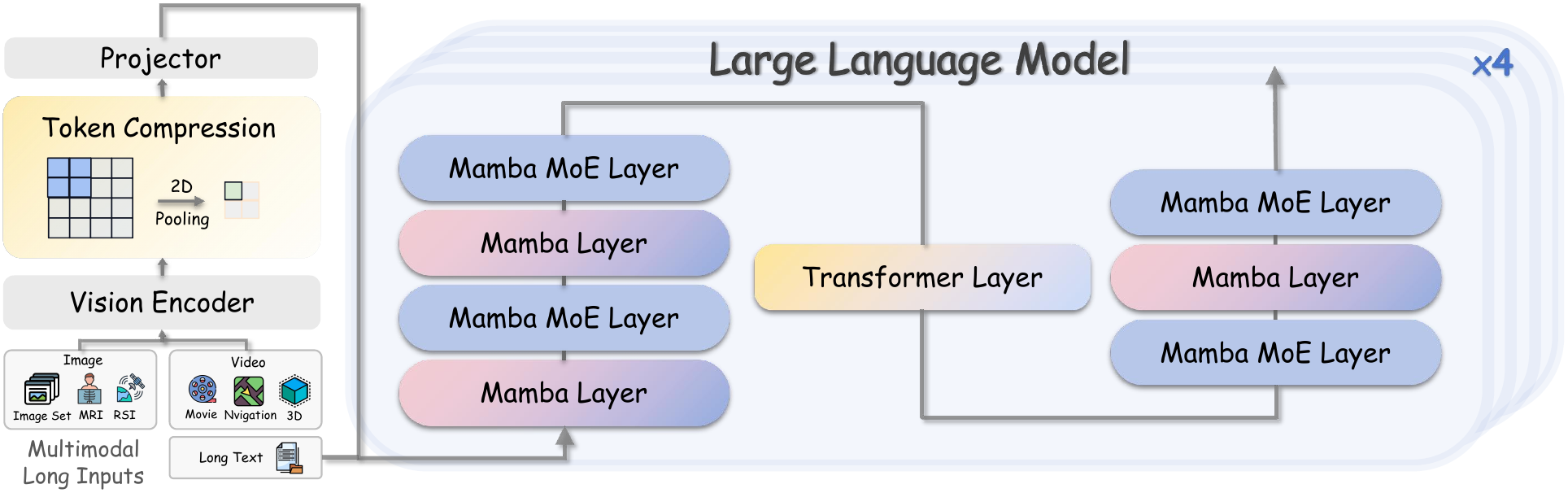}
% \vspace{-1mm}
\caption{\textbf{Architecture of LongLLaVA.} The LongLLaVA model is capable of (1) accommodating a variety of multimodal inputs and efficiently processing image tokens via 2D token compression; (2) uniformly managing the preprocessed inputs within its hybrid LLM architecture.}
% \vspace{-2mm}
\label{fig:arch}
\end{figure*}

\subsection{Implementation for Hybrid Architecture}
% \vspace{-2mm}
\begin{table}[h]\footnotesize
% \addtolength\tabcolsep{-1pt}
\centering
\begin{tabular}{@{}lccccc@{}}
\toprule
Arch & HellaSwag & NQ & BoolQ & ARC-C \\ \midrule
Attention (1:0)  & 62.4 & 14.5 & 60.9 & 34.6 \\
Hybrid (1:3)  & 65.1 & 16.5 & 60.6 & 36.8 \\
\rowcolor{gray!15} Hybrid (1:7) & 65.1 & 16.0 & 64.4 & 34.8 \\
Mamba (0:1) & 62.6 & 14.5 & 61.1 & 34.1 \\ \bottomrule
\end{tabular}
% \vspace{-2mm}
\caption{Performance comparison of different hybrid architecture ratios on a 1.3B parameter model trained with 250B tokens. Details provided in Appendix~\ref{sec:ratio}.}
% \vspace{-2mm}
\label{tab:arch_comparison}
\end{table}

Our hybrid architecture leverages established foundation model research. Its Mixture of Experts (MoE) configuration adopts the layer-wise pattern proposed by Jamba \citep{lieber2024jambahybridtransformermambalanguage}, with expert layers integrated every two layers. For the Attention-Mamba blend ratio, previous work \citep{wang2024mamballamadistillingaccelerating} evaluated ratios such as 1:0, 1:1, 1:3, and 1:7, and found substantial performance gains when transitioning from pure Mamba (0:1 ratio) to a 1:7 blend, with diminishing returns as the transformer proportion increases further. This conclusion is further supported by \citet{lieber2024jambahybridtransformermambalanguage}. Experiments on 1.3B parameter architectures trained on 250 billion tokens, with results presented in Table~\ref{tab:arch_comparison} and details provided in Appendix~\ref{sec:ratio}, show only a marginal performance difference between the 1:7 and 1:3 ratios. Crucially, the 1:3 configuration is also significantly more computationally expensive. Thus, balancing empirical performance and computational efficiency, we selected the 1:7 configuration as optimal.

\section{LongLLaVA}
% \vspace{-3mm}

To address the aforementioned challenges and enhance the model's adaptability to long-context, multi-image scenarios, we introduce improvements from three perspectives: \textit{multi-modal model architecture} (Sec.~\ref{sec:arch}), \textit{data processing protocol} (Sec.~\ref{sec:data}), and \textit{training strategy} (Sec.~\ref{sec:train}).

% \vspace{-3mm}
\subsection{Multi-modal Architecture}
\label{sec:arch}
% \vspace{-1mm}

The architecture consists of three core components inspired by LLaVA~\citep{li2024llava}: the Vision Encoder, the Projector, and the LLM.
% The primary strategies for adapting to multimodal long-context are predominantly derived from two aspects.

\paragraph{Vision Information Processing}
We employ CLIP\footnote{\texttt{openai/clip-vit-base-patch32}} as the vision encoder to encode visual information and a two-layer MLP as the projector to map vision features into the text embedding space suitable for the LLM. Prior to projection, bilinear pooling is applied, reducing the token representation of an image from 576 to 144 by aggregating $2\times2$ patch units into a single token. This approach effectively conserves training and inference time while maintaining essential spatial relationships between patches. In Section~\ref{sec:abl}, we further discuss the impact of this token reduction on performance and explore strategies for its mitigation.

\paragraph{LLM Architecture}
Our model employs a hybrid LLM architecture comprising four stacks of hybrid layers, each integrates Transformer and Mamba layers in a 1:7 ratio, as depicted in Figure~\ref{fig:arch}. It also features a Mixture of Experts (MoE) approach in every other layer, utilizing 16 experts and selecting the top-2 experts for each token. RMSNorm~\citep{zhang2019rootmeansquarelayer} is used between layers to enhance normalization, although positional embeddings are omitted. The model incorporates Grouped Query Attention (GQA)~\citep{ainslie2023gqatraininggeneralizedmultiquery} and SwiGLU activation functions~\citep{shazeer2020gluvariantsimprovetransformer}, similar to other large language models. 
The total parameter count of the model is \texttt{53B}, with activation parameters during inference totaling \texttt{13B}; we designate this model as \textbf{LongLLaVA-A13B}.
In an effort to make the model more efficient, we have retained only the \texttt{Expert-0} in the \texttt{Mamba MoE Layer}\footnote{\revise{We chose Expert-0 due to minimal performance differences, detailed in Appendix~\ref{app:expert}.}}, thereby constructing \textbf{LongLLaVA-9B}.

% For long-context training, we adopted \textit{LenCat}, a dynamic sampling packing strategy based on length priority. In addition, in order to fully explore the performance of long context for single-image understanding, we adopted \textit{BestFit}, an image segmentation technique that does not perform image resolution scaling and overlapping segmentation.

% By integrating Transformer, Mamba, and MoE components, Jamba flexibly balances the sometimes conflicting goals of low memory usage, high throughput, and high quality. For the Key-Value (KV) cache, Jamba achieves an 8-fold reduction in size compared to the vanilla Transformer. In terms of throughput, Jamba demonstrates greater computational efficiency, especially when handling long attention lengths, which consume a significant portion of computational resources.

% \vspace{-3mm}

\begin{figure*}[t]
%\vspace{-14mm}
\begin{AIbox}{\footnotesize{Data Processing Protocol}}\footnotesize
\textbf{In the Following Statement: \space \texttt{<Image>=<img><img\_token>...</img>}}\\
\textit{For Single-image}: \space “\texttt{<Image>\textbackslash n} What is this?” \\
\textit{For Multi-image}: \space “\texttt{<Image>\textbackslash n} This is a cat. \texttt{<Image>\textbackslash n}This is a:” \\ 
\textit{For Video}:  \space “\texttt{<vid><Image><t>...<Image></vid>\textbackslash n} What are they?” \\ 
\textit{For Patched-image}: \space “\texttt{<Image>\textbackslash n\texttt{<Image>..\textbackslash n..<Image>}\textbackslash n} What are they?” 
\end{AIbox} 
% \vspace{-3mm}
\caption{\textbf{Data Processing Protocol for LongLLaVA}.We utilized different tokens to distinguish various modal information, and to identify the spatial and temporal relationships within images.}
% \vspace{-1mm}
\label{fig:pro} 
\end{figure*}

\begin{figure*}[t]
    %\vspace{-3mm}
    \centering
    \includegraphics[width=0.8\linewidth]{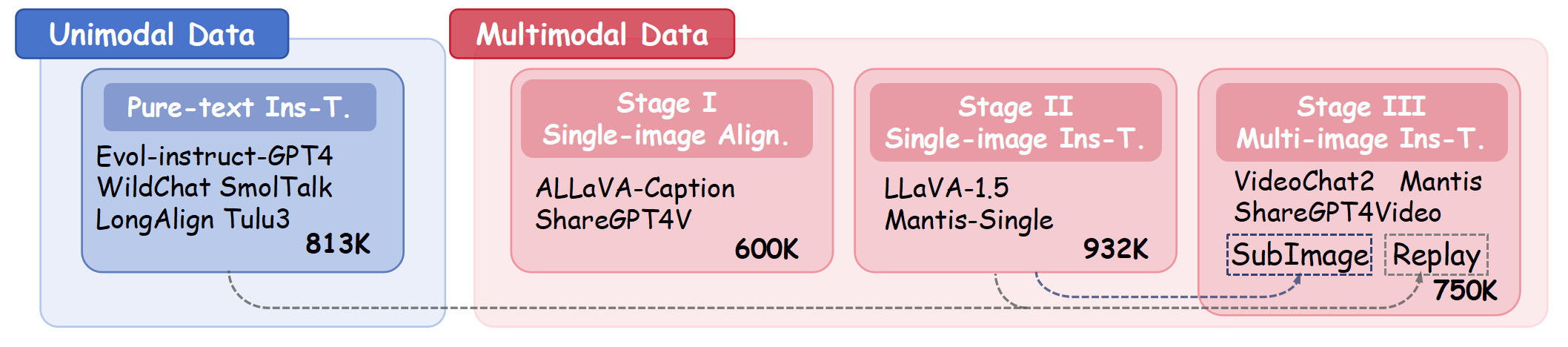}
    % \vspace{-3mm}
    \caption{\textbf{Dataset Taxonomy of LongLLaVA}. \texttt{Replay} refers to data sampled from former phase to maintain single-image and dialogue understanding ability. \texttt{SubImage} denotes a constructed dataset for understanding complex single images divided into sub-images. Ins-T. and Align. refer to instruction-tuning and alignment, respectively.}
    \label{fig:data}
    % \vspace{-1mm}
\end{figure*}

%\vspace{-3mm}
% \vspace{-1mm}

% \subsection{Data Processing Protocol} 
% \label{sec:data}
% To ensure that the model effectively distinguishes between temporal and spatial dependencies among images in multi-image scenarios and performs well across various tasks, we meticulously differentiated special characters in different scenarios. As shown in Figure~\ref{fig:pro}, these special characters comprehensively address the various relationships between images in different contexts, thereby enhancing the model's adaptability to diverse tasks.

% \textbf{Regular Single and Multiple Images.} For this type of inputs, we use \texttt{<img>} and \texttt{</img>} to enclose image tokens, helping the model differentiate between image and text tokens.

% \textbf{Video.} For video inputs, to enable the model to understand the temporal relationship between frames, we first use \texttt{<vid>} and \texttt{</vid>} to enclose image tokens. Additionally, we add the special symbol \texttt{<t>} between different frames to represent the temporal dependency between them.

% \textbf{High Resolution Image.} For complex image understanding that require dividing an image into multiple sub-images, we use \texttt{\textbackslash n} to separate the main image from its sub-images. For the arrangement of sub-images, we traverse from the top-left to the bottom-right, adding \texttt{\textbackslash n} between split lines to preserve the spatial positions of the sub-images.

\subsection{Data Processing Protocol}
\label{sec:data}
To ensure the model can effectively distinguish temporal from spatial dependencies in multi-image inputs and perform robustly across diverse tasks, we have meticulously designed and differentiated special tokens for various scenarios. As illustrated in Figure~\ref{fig:pro}, these tokens are tailored to represent the complex relationships between images in varying contexts, thereby enhancing the model's adaptability to a wide range of tasks.

\paragraph{Regular Single and Multiple Images} For regular single and multiple image inputs, we use \texttt{<img>} and \texttt{</img>} tokens to demarcate image-derived token sequences. This helps the model to differentiate these from textual tokens in the input stream.

\paragraph{Video} For video inputs, to enable the model to comprehend the temporal relationships between frames, we enclose the entire sequence of frame tokens with \texttt{<vid>} and \texttt{</vid>}. Furthermore, the special token \texttt{<t>} is inserted between consecutive frames to signal their temporal dependency.

\paragraph{High Resolution Image} For scenarios involving complex image understanding, such as high-resolution images segmented into multiple sub-images, we utilize the \texttt{\textbackslash n} token for structural organization. This token is first used to separate the representation of the global image from the block of its constituent sub-images. Additionally, when arranging these sub-images, which are typically ordered in a top-left to bottom-right raster scan, \texttt{\textbackslash n} is inserted between the rows of sub-images. This approach preserves their relative spatial positions within the linearized input.

\begin{table*}[t]\footnotesize
% \vspace{-2mm}
\small
\centering
\begin{threeparttable}
\addtolength\tabcolsep{-3.8pt} 
\begin{tabular}{lcc|cccc|cccc|cc}
\toprule
\multicolumn{1}{c}{\multirow{2}{*}{Model}} &\multicolumn{1}{c}{\multirow{2}{*}{PFLOPs}} &\multicolumn{1}{c|}{\multirow{2}{*}{\revise{\#P.}}} & \multicolumn{4}{c|}{MileBench}  & \multicolumn{4}{c|}{Video-MME w/o subs} & \multicolumn{1}{c}{\multirow{2}{*}{MVBench}} & \multicolumn{1}{c}{\multirow{2}{*}{LongVideo*}}\\
\multicolumn{1}{c}{} &\multicolumn{1}{c}{} &\multicolumn{1}{c|}{}  & Temporal & Semantic & IR & Avg. & Short    & Medium    & Long   & Avg. & \multicolumn{1}{c}{} & \multicolumn{1}{c}{} \\ \midrule
\rowcolor{gray!15}\multicolumn{13}{c}{\textbf{Proprietary Models}} \\
GPT-4V   &- & - & 45.6& 58.9& 86.7  & 63.7 &70.5& 55.8 & 53.5   & 59.9  & 43.5 & 59.1\\
GPT-4o   &- & -& \textbf{56.2}& \textbf{63.5}& \textbf{88.8}  & \textbf{69.5} &72.5& 63.1 & 58.6   & 64.7  & - & 66.7 \\
Gemini-1.5-Pro &- & -& 50.2& 58.3& 88.0  & 65.5 &\textbf{78.8}& \textbf{68.8} & \textbf{61.1}   & \textbf{69.6}  & - & 64.0 \\
Claude3-Opus &- & -& 37.4& 48.1& 25.0  & 36.8 & 70.5& 57.4 & 51.2   & 59.7  & - & - \\
\midrule
\rowcolor{gray!15}\multicolumn{13}{c}{\textbf{Open-source MLLMs}} \\ 
 LongVA&4.90 & 8B & - & - & - & - & 61.1& 50.4& 46.2& 52.6 & - & -\\
InternVL2 & 4.10 & 8B & - & - & - & - & - & - & - & 56.3 & 65.8 & 54.6 \\
InternVL2.5 & 4.10 & 8B & - & - & - & - & - & - & - & 64.2 & \textbf{72.0} & \textbf{60.0} \\
OmChat &3.90 & 8B & 51.4 & 52.0 & 34.2 & 45.9 &- & - & - & -  & 50.2 & -\\
LongVILA&3.90  & 8B & - & - & - & - &61.8 & 49.7& 39.7 & 50.5  & - & - \\
Qwen2-VL & 3.80 & 7B & - & - & - & - & - & - & - & 63.3 & 67.0 & - \\ 
Qwen2.5-VL & - & 7B & - & - & - & - & - & - & - & 65.1 & 69.6 & 56.0 \\
\midrule
\rowcolor{gray!15}\multicolumn{13}{c}{\textbf{Open-source Efficient MLLMs}} \\
VideoLLaMA2 &3.71 & 7B & 3.2 & 6.6 & 4.5 & 5.0 & 55.9& 45.4 & 42.1   & 47.8  & 34.1 & 40.3 \\
mPLUG-Owl3 & - & 8B & - & - & - & - & - & - & - & 53.5 & 54.5 & 52.1 \\
Phi-3-Vision &2.68 & 4B & 46.9 & 50.0 & 18.7 & 38.5 &- & - & - & -  & - & 49.6\\
Cobra & 1.02 & 7B & - & - & - & - & - & - & - & 49.5 & - & - \\
VideoChat2 &0.24 & 7B & 25.5& 25.5& 9.2   & 20.1 &48.3& 37.0 & 33.2   & 39.5  & 51.9 & 39.3 \\
% \cmidrule(l{1.0em}r{1.0em}){1-13}
\rowcolor{mediumgray}\textbf{LongLLaVA-9B} & \textbf{0.15} & 9B & 54.2 & 52.4 & 53.2 & 53.2 & 59.6 & 50.3& 42.7 & 50.9 & 59.4 & 51.9 \\
\rowcolor{mediumgray}\textbf{LongLLaVA-A13B} & \textbf{0.22} & 53B & \textbf{56.2} & \textbf{58.6} & \textbf{68.5} & \textbf{59.2} & \textbf{62.9} & \textbf{52.2} & \textbf{46.4} & \textbf{53.8} & \textbf{64.6} & \textbf{53.5} \\ \bottomrule
\end{tabular}
% \begin{tablenotes}
% \scriptsize
% \item[*] `More Data' indicates model trained for around two epochs (82 hours $\times$ 8 $\times$ A800-80G) to ensure comparable training time to LongVA (84 hours $\times$ 8 $\times$ A100-80G).
% \end{tablenotes}
\end{threeparttable}
% \vspace{-3mm}
\caption{Multi-image Evaluation Results: PFLOPs indicate floating-point operations per 128 images. LongVideo* abbreviates LongVideoBench. All evaluations used FP16 precision.} 
\label{tab:multi_image} 
% \vspace{-4mm}
\end{table*}

\subsection{Training Strategy}
\label{sec:train}
%\vspace{-3mm}

Our training strategy employs both single-modal and multi-modal adaptations to transform a pre-trained language model into a multimodal long-context model.

\paragraph{Pure-text Instruction Tuning}
Initially, we enhance the pre-trained language model's capacity to follow instructions of varying lengths within pure-text contexts. This is accomplished using a comprehensive dataset of 813k pure-text entries, aggregated from Evol-instruct-GPT4~\citep{xu2023wizardlmempoweringlargelanguage}, WildChat~\citep{zhao2024wildchat1mchatgptinteraction}, SmolTalk~\citep{allal2025smollm2smolgoesbig}, and high-quality data sampled from Tulu3~\citep{lambert2025tulu3pushingfrontiers} via DEITA~\citep{liu2024what}, alongside LongAlign~\citep{bai2024longalignrecipelongcontext}.

For multi-modal adaptation, we adopt a progressive training approach, which offers better variable control and increases model performance~\citep{fu2024data}. Building upon the \textit{Single-image Alignment} and \textit{Single-image Instruction-tuning} stages outlined in LLaVA~\citep{li2024llava}, we introduce a \textit{Multi-image Instruction-tuning} stage to systematically enhance the model's long-context capabilities. Details of dataset usage are provided in Figure~\ref{fig:data}.

\paragraph{Stage I: Single-image Alignment}
This initial multi-modal stage aims to align visual features with the textual modality. We utilize datasets such as ALLaVA-Caption~\citep{chen2024allava} and ShareGPT4V~\citep{chen2023sharegpt4vimprovinglargemultimodal}, collectively comprising approximately 600K high-quality image-caption pairs. During this phase, only the projector is trained, while the parameters of the Visual Encoder and the LLM remain frozen.

\paragraph{Stage II: Single-image Instruction Tuning}
The objective of this stage is to imbue the model with multimodal instruction-following capabilities. We employ datasets including LLaVA-1.5~\citep{liu2023llava} and Mantis-Single~\citep{jiang2024mantisinterleavedmultiimageinstruction}, totaling 932K high-quality question-answer pairs. Only the Visual Encoder's parameters are frozen. 

\paragraph{Stage III: Multi-image Instruction Tuning} This stage fine-tunes the model for instruction following in multi-image scenarios. Training data includes 200K instances from Mantis~\citep{jiang2024mantisinterleavedmultiimageinstruction}, 200K from VideoChat2~\citep{li2024mvbenchcomprehensivemultimodalvideo}, and 50K from ShareGPT4Video~\citep{chen2024sharegpt4video}. The \texttt{Replay} component, incorporating 200K single-image and 50K pure-text instruction-tuning instances, preserves established single-image comprehension and pure-text dialogue capabilities. Furthermore, the \texttt{Sub-Image} component enhances the interpretation of complex single images processed as segments; this is formed using 50K single-image instruction instances where original images are padded and segmented into sub-images of size $336\times336$.

\begin{table*}[t]\footnotesize
% \vspace{-2mm}
\centering
\addtolength\tabcolsep{-1.7pt} 
\begin{tabular}{llr|rrr|rrr|rrr|r}
\toprule
\multicolumn{1}{c}{\multirow{2}{*}{Video MLLM}} & \multicolumn{1}{c}{\multirow{2}{*}{PFLOPs}} & \multicolumn{1}{c|}{\multirow{2}{*}{\revise{\#P}}} & \multicolumn{3}{c|}{Retrieval} & \multicolumn{3}{c|}{Ordering}  & \multicolumn{3}{c|}{Counting}  & \multirow{2}{*}{Avg.} \\
\multicolumn{1}{c}{}  & \multicolumn{1}{c}{} & \multicolumn{1}{c|}{}  & E    & I-1  & \multicolumn{1}{l|}{I-2}  & E    & I-1  & \multicolumn{1}{l|}{I-2}  & E-1  & E-2  & \multicolumn{1}{l|}{I}    &   \\ \midrule
% \rowcolor{gray!15}\multicolumn{13}{c}{\textbf{Proprietary Models}} \\
% {Gemini-1.5} & - & - & 100.0  & 96.0   & {76.0}   & 90.7 & 95.3 & {32.7} & 60.7 & 7.3  & {42.0}   & 66.7   \\
{GPT-4o} & - & - & 100.0  & 98.0   & {87.3} & 88.4 & 86.6 & {45.2} & 36.8 & 0.0    & {36.1} & 64.4   \\
{GPT-4V} & - & - & 100.0  & 99.3 & {82.0}   & 42.6 & 22.8 & {23.0}   & 37.6 & 0.0    & {32.4} & 48.9   \\ \midrule
\rowcolor{gray!15}\multicolumn{13}{c}{\textbf{Open-source MLLMs}} \\ 
% {VideoChatGPT} & 4.7  & 4.7  & {0.7}  & 2.7  & 11.3 & {0}    & 2    & 4    & {6.7}  & 4.1    \\
Qwen2-VL & 0.87 & 7B  & 98.0  & \textbf{76.0}  & 33.3 & 16.0 & 12.7 & 8.7 & 26.0 & 9.3  & 24.7 & 33.9  \\
VideoLLaMA2 & 0.85 & 7B & 1.2  & 26.0   & {6.0}    & 0.0    & 0.0    & {0.0}    & 2.0    & 4.7  & {0.7}  & 4.5    \\
% {LLaMA-VID}    & 28   & 28   & {25.1} & 0    & 0    & {0}    & 2.4  & 14.7 & {7.1}  & 11\\
% {Video-LLAVA}  & 26   & 17.3 & {23.8} & 0.7  & 0.7  & {0}    & 1.1  & 16.7 & {2}    & 8.2    \\
% {VideoChat2}  & 0.08 & 7B & 43.4 & 40.0   & {14.6} & 0.0    & 0.0    & {1.3}  & 4.4  & 8.0    & {12.4} & 12.4   \\
% {LLAVA-N-Video} & 32B & 56.7 & 56.7 & {44.2} & 0.0    & 0.0    & {0.0}    & 14.6 & 25.3 & {15.5} & 20.1   \\
% {ST-LLM}  & 58   & 64.7 & {31.3} & 2    & 0.7  & {0}    & 21.3 & 17.3 & {16.7} & 22.7   \\ 
\textbf{LongLLaVA-9B} & 0.07 & 9B&  98.3  & 57.2 & 96.3 & 24.2 & \textbf{57.2} & 24.3  & 24.5 & 21.0 & 26.0 & 44.4  \\ 
\textbf{LongLLaVA-A13} & 0.09 & 53B & \textbf{100}  & 73.3 & {\textbf{100.0}} & \textbf{37.5}    & 35.3 & {\textbf{34.8}}    & \textbf{36.0} & \textbf{23.7 }& {\textbf{28.0}} & \textbf{52.1}  \\ 
\bottomrule
\end{tabular}
% \vspace{-2mm}
\caption{Long Context MLLMs' Atomic Capabilities Analysis using VNBench~\citep{zhao2024needlevideohaystackscalable}. PFLOPs refers to the number of floating-point operations required for inference on 54 images, which corresponds to the average number of frames extracted from the dataset videos at 1 FPS.} 
% \vspace{-1mm}
\label{tab:ana1} 
\end{table*}

% \vspace{-1mm}
\section{Experiments}
\label{sec:eva}
% \vspace{-1mm}

\subsection{Experimental Setup} 

To manage large-scale, diverse datasets during training, data items are randomly sampled and concatenated into sequences of 176K tokens, with individual items separated by the \texttt{<eos>} token. The model is trained for a single epoch on a distributed setup of $3\times8$ A800 GPUs. A cosine learning rate schedule is employed, with a $0.03$ warm-up proportion and a peak learning rate of \texttt{1e-5}. Detailed information on multi-image evaluation benchmarks and baselines is available in Appendix~\ref{app:eval}. Unless otherwise noted, both \textit{LongLLaVA-9B} and \textit{LongLLaVA-A13B} models are evaluated using \texttt{Int8} quantization to reduce computational costs while maintaining performance. Hereafter, \texttt{LongLLaVA} will refer to the \textit{LongLLaVA-A13B} model. Information regarding the evaluation of fundamental single-image understanding capabilities is provided in Appendix~\ref{app:single_eval}.

\vspace{-1mm}
\subsection{Results}
\label{sec:main_result}
% \vspace{-2mm}

\textbf{Main Results} As detailed in Table~\ref{tab:multi_image}, LongLLaVA exhibits strong performance among open-source models on the MileBench benchmark. It also demonstrates notable results in retrieval-oriented tasks, indicating its proficiency in processing multi-image inputs. Furthermore, its effectiveness is reflected in its performance on video benchmarks such as Video-MME~\citep{fu2024video} and MVBench~\citep{li2024mvbenchcomprehensivemultimodalvideo}. A key aspect is its achievement of these results with a substantially lower computational cost, specifically an order of magnitude fewer FLOPs. This approach, therefore, presents a balance of enhanced performance relative to other architecture optimization methods while maintaining considerable operational efficiency in comparison to several SOTA models.

% Remarkably, despite achieving these impressive results, LongLLaVA operates with an order of magnitude fewer FLOPs compared to other models. This efficiency in computational resources not only underscores LongLLaVA's advanced performance but also its optimization in resource management. These results reflect a significant advancement in the research community's efforts to close the performance gap with commercial models.

%\vspace{-3mm}
% \subsection{Diagnostic Evaluation of Long-Context MLLMs}
% : Retrieval, Ordering and Counting
%\vspace{-3mm}

\begin{figure}[t]
    \centering
    % \vspace{-5mm}
    \includegraphics[width=1\linewidth]{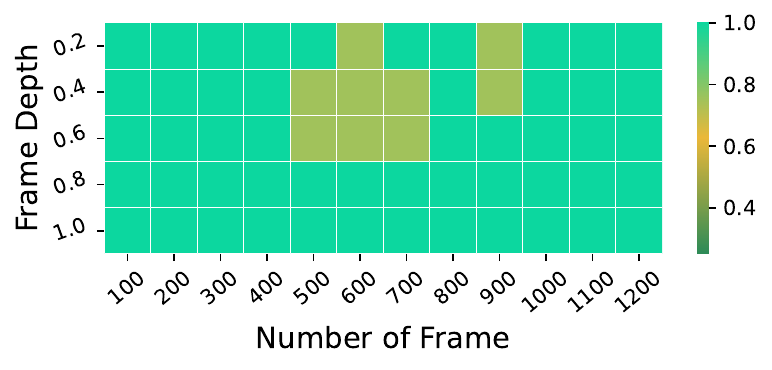}
    % \vspace{-8mm}
    \caption{Video-NIAH~\citep{zhang2024longcontexttransferlanguage} evaluated on one A800 80GB GPU.}
    % \vspace{-2mm}
    \label{fig:map}
\end{figure}

\paragraph{Diagnostic Evaluation of Long-Context} To address limitations in evaluating long-context MLLMs, we conducted a diagnostic assessment using VNBench~\citep{zhao2024needlevideohaystackscalable}, a synthetic video framework evaluating atomic capabilities like retrieval, ordering, and counting. As detailed in Table~\ref{tab:ana1}, LongLLaVA's performance is on par with, and sometimes exceeds, leading closed-source models such as GPT-4V, while also outperforming other open-source counterparts in managing extensive contexts. Further substantiating its retrieval strength, LongLLaVA also achieves nearly 100\% accuracy on the 1200-image V-NIAH evaluation framework~\citep{zhang2024longcontexttransferlanguage} without additional training, as depicted in Figure~\ref{fig:map}. These findings collectively indicate LongLLaVA's significant proficiency in long-context understanding and information retrieval.

\subsection{Ablation Study}
\label{sec:abl}

\begin{table}[t]\tiny
    \centering
    % \vspace{-2mm}
    \small % Use \scriptsize for even smaller text if necessary
    \addtolength\tabcolsep{-4.5pt}
    \begin{tabular}{lccccccc}
    \toprule
    Model & MMLU & \multicolumn{1}{c|}{BBH} & GQA & MMMU & SEED$^{v1}_{img}$ & Mile \\
    \midrule
    & & \multicolumn{1}{c|}{} & \multicolumn{4}{c}{\cellcolor{gray!15}\textbf{With LLaVA-1.5 Recipe}} \\ 
    Vicuna-13B & \textbf{55.3} & \multicolumn{1}{c|}{\textbf{40.5}} &\textbf{63.3} & 34.4 & 68.2 & 27.6 \\
    \textbf{Jamba-9B} & 54.3 & \multicolumn{1}{c|}{38.4} & 62.3 & \textbf{36.2} & \textbf{70.1} & \textbf{28.2} \\
    % % \midrule
    % Difference & -1.0 & \multicolumn{1}{c|}{-2.1} & -1.0 & +1.8 & +1.9 & +0.6 \\
    \bottomrule
    \end{tabular}
    % \vspace{-2mm}
    \caption{Ablation on MLLM Backbone Architectures.}
    % \vspace{-4mm}
    \label{tab:hybrid-architecture}
\end{table}

\begin{table}[t]
% \vspace{-4mm}
\centering
% \begin{table}[H]
\small
\addtolength\tabcolsep{-4.5pt} 
        \footnotesize
\begin{tabular}{>{\raggedright\arraybackslash}p{20.9mm}lccccc}
\toprule
 Method & \#T & GQA & MMMU & SQA& SEED$^{v1}_{img}$ & Mile\\ \midrule
 % \rowcolor{gray!15}\multicolumn{7}{c}{\textbf{Architecture \& Data Abalation on LongLLaVA-A13B}}\\
 % LLaVA-1.5-13B& 576 & 63.3& 34.4& 71.6& 68.2&27.6\\
\rowcolor{gray!15} \multicolumn{7}{c}{\textbf{Ablation on Token Compression}} \\
 \textbf{Jamba} & 576 & 63.2 & 41.4 & 75.4 & 69.8 & 38.2 \\
  1D Pooling & 144 & 60.4 & 42.0 & 73.9 & 66.3 & 36.2  \\
 \textbf{2D Pooling} & 144 & 61.3 & 42.1 & 75.2 & 67.4 & 37.7 \\
% \cmidrule(l{1.0em}r{1.0em}){1-7}
 \rowcolor{gray!15} \multicolumn{7}{c}{\textbf{Ablation on Dataset Construction}} \\
\textbf{+S-image Data} & 144 & 62.2 & 42.1 & 75.9 & 68.9 & 50.0 \\
\textbf{+M-image Data} & 144 & 59.9 & 39.2 & 73.4 & 65.3 & 57.4\\
 \midrule
% % Training Strategy & \\
 % \rowcolor{gray!15}\multicolumn{7}{c}{\revise{\textbf{Training Strategy Abalation on LongLLaVA-9B}}}\\
 \rowcolor{gray!15} \multicolumn{7}{c}{\textbf{Ablation on Training Strategies}} \\
 Stage1\&2\&3& 144 & 56.9& 32.8& 67.2& 66.9& 42.2\\
Stage1, 2\&3& 144 & 57.6& 33.2& 70.2& 68.4&44.2\\
\textbf{Stage1, 2, 3}& 144 & 58.4& 34.4& 69.9& 67.9& 46.5\\
\bottomrule
\end{tabular}
% \vspace{-2mm}
\caption{Ablation on token compression, dataset construction and training strategies. 1D and 2D denote different pooling strategies. \#T: the token count for one image. \&: the combination of the stages. S-image: single-image. M-image: multiple-image.} 
\label{tab:ab1} 
% \vspace{-6mm}
\end{table}

\paragraph{Ablation on MLLM Backbone Architectures}
To assess the impact of hybrid architectures on MLLM performance, we use Vicuna-13B~\citep{vicuna2023} and Jamba-9B~(trained as described in Appendix~\ref{app:expert}) as initial LLMs. As shown in Table~\ref{tab:hybrid-architecture}, both models perform similarly before multimodal adaptation, with Vicuna-13B slightly ahead, ensuring a fair comparison. After training with the LLaVA-1.5 training recipe~\citep{liu2024llavanext}, the hybrid architecture consistently achieves better results on most multimodal benchmarks, despite slightly lower base LLM performance. This demonstrates that hybrid architecture is efficient and has no adverse effect on the multimodal adaptation.

\paragraph{Ablation on other Methods}
Ablation results for other methods are presented in Table~\ref{tab:ab1}. For \textbf{token compression}, pooling significantly reduces computational cost while keeping performance degradation within acceptable limits. Moreover, two-dimensional pooling with a $12\times12$ label arrangement offers clear advantages over one-dimensional pooling. Regarding \textbf{dataset construction}, the quality of our single-image training data surpasses that of LLaVA-1.5, and incorporating multi-image data substantially improves the model's performance on multi-image tasks. In terms of \textbf{training strategies}, progressive training is more effective than mix-training for multi-image tasks, while maintaining comparable results on single-image tasks. Due to space constraints, ablation results for replay data are provided in Appendix~\ref{app:abl-2}.

% \vspace{-3mm}
\section{Analysis}
% \vspace{-1mm}
\subsection{Scaling Law of Image Numbers}
% \vspace{-1mm}

Processing more images enables models to handle additional video frames and provide more examples for few-shot learning. To investigate the effects of increasing the number of frames and examples, we evaluate LongLLaVA on the Video-MME~\citep{fu2024video} and LongLLaVA-9B on the VL-ICL~\citep{zong2024vliclbenchdevildetails}, respectively.

\begin{figure}[t]
\centering
\includegraphics[width=0.95\linewidth]{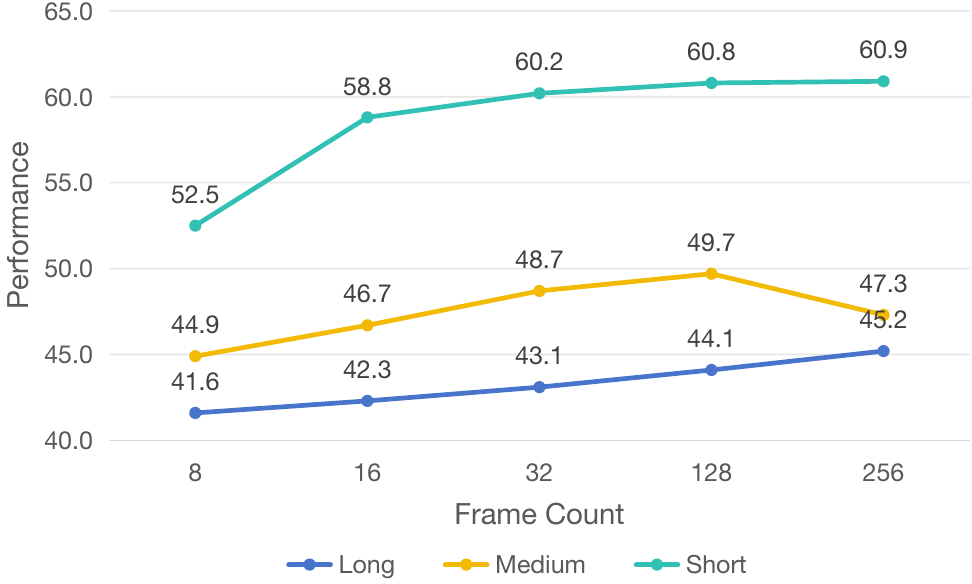}
% \vspace{-3mm}
\caption{Performance on the Video-MME benchmark as the number of sampled frames per video increases.}
\label{fig:videomme}
\end{figure}

\begin{figure}[t]
\centering
% \vspace{-1mm}
\includegraphics[width=0.42\textwidth]{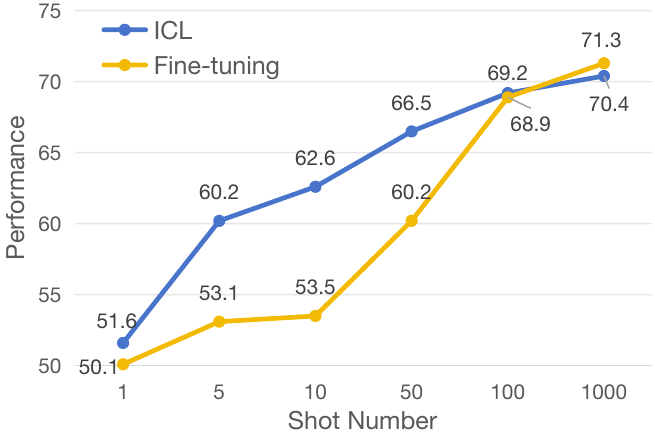}
% \vspace{-2.5mm}
\caption{Performance comparison between Many-Shot ICL and fine-tuning on VL-ICL.}
\label{fig:app3}
% \vspace{-6mm}
\end{figure}

\paragraph{Scaling Number of Frames}
Video-MME evaluates a model’s ability to extract information from videos. As shown in Figure~\ref{fig:videomme}, increasing the number of sampled frames steadily improves performance, peaking at 256 frames. This indicates that the model effectively utilizes additional visual information from more frames.

\paragraph{Scaling Number of Shots} Fine-tuning LLMs can be costly and impractical, especially with limited data or frequently changing tasks. In contrast, many-shot in-context learning (ICL) allows models to utilize more task-specific examples during inference without retraining~\citep{agarwal2024manyshotincontextlearning}. To evaluate this, we compare performance across different shot numbers and fine-tuning on the “Matching Image” task from VL-ICL, where each input is an image pair $x = \{x_1, x_2\}$ and the output $y$ indicates if a predefined relation $r$ holds. As shown in Figure~\ref{fig:app3}, ICL outperforms fine-tuning up to around 100 shots; however, when the number of examples exceeds 1,000, fine-tuning becomes more effective.

\begin{figure}[t]
% \vspace{-1mm}
\centering
\includegraphics[width=1.0\linewidth]{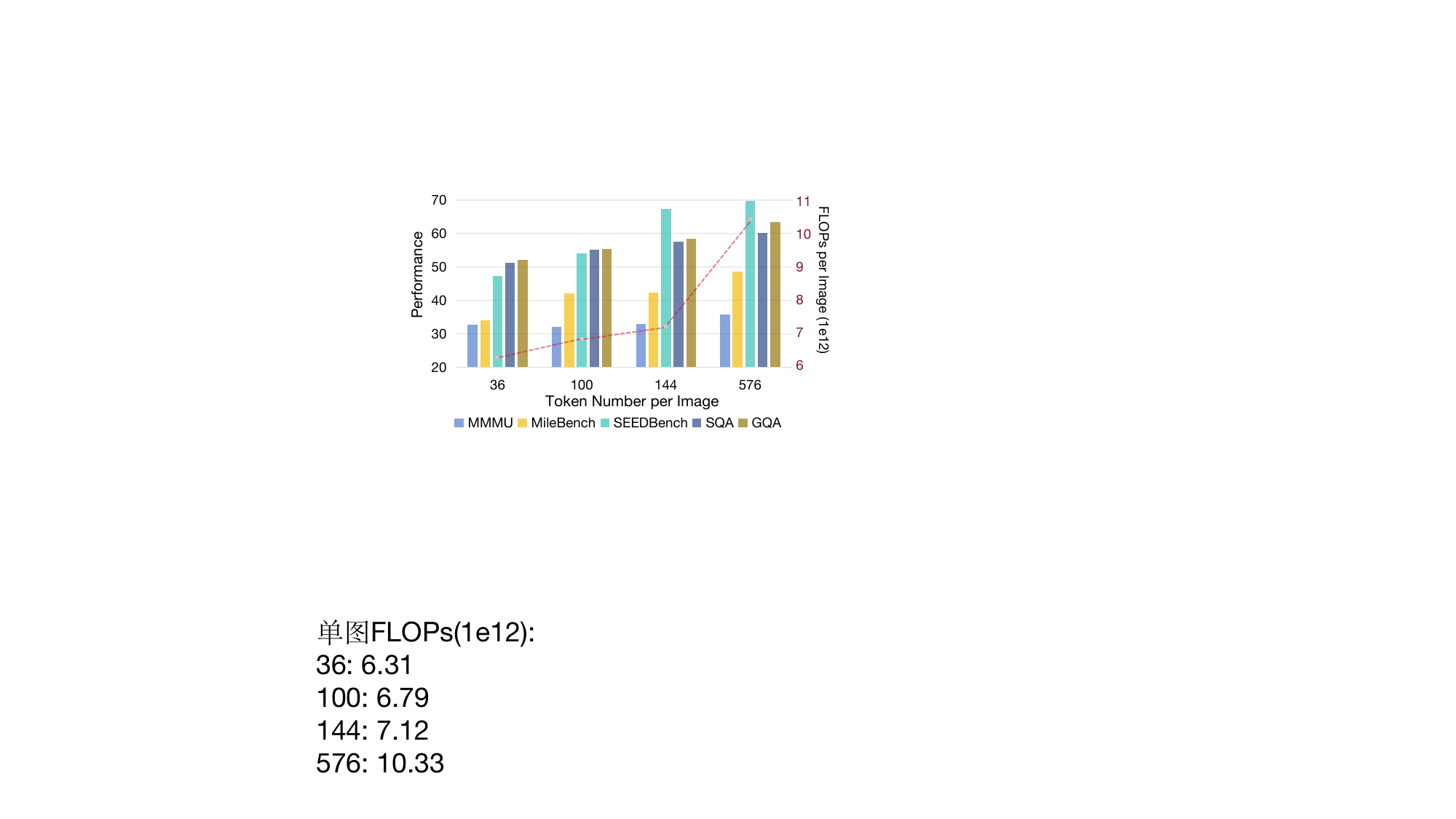}
% \vspace{-5.5mm}
\caption{Performance and inference cost across five benchmarks with varying number of tokens per image.}
\label{fig:token_number}
% \vspace{-1mm}
\end{figure}

\begin{figure}[t]
\centering
\includegraphics[width=0.99\linewidth]{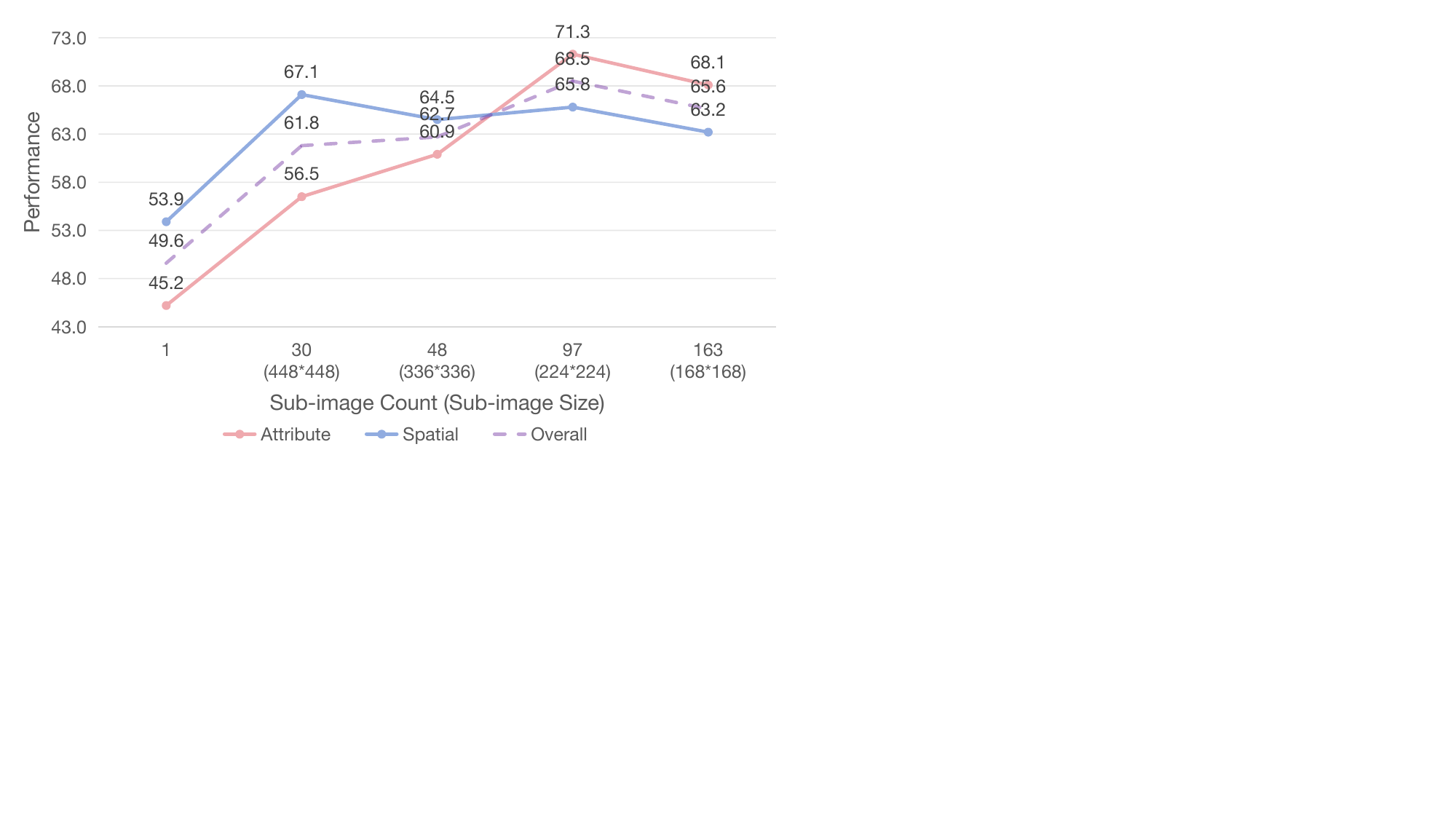}
% \vspace{-5.5mm}
\caption{Performance on V* with different Sub-Image counts as Mitigating Token Compression Strategy.}
% \vspace{-5mm}
\label{fig:v-star}
\end{figure}

\subsection{Impact and Mitigation of Token Compression}
\label{sec:token}

To quantify the impact of token compression on visual understanding and to evaluate corresponding mitigation strategies, we conduct a series of experiments. Our analysis leverages five general vision-language benchmarks alongside V* Bench~\citep{vstar}, a specialized benchmark designed to assess the localization of small objects within large images, a task known to be particularly sensitive to information loss.

\paragraph{Impact of Token Compression}
As shown in Figure~\ref{fig:token_number}, setting the token count to 144 per image substantially reduces inference cost while incurring minimal degradation in overall performance. This effective trade-off is particularly evident on SEEDBench, demonstrating that a significant reduction in computational overhead is achievable without compromising the model's general capabilities.

\paragraph{Mitigation through Image Partitioning}
To counteract the information loss inherent in token compression, we find that a simple strategy of partitioning the input image is highly effective for fine-grained tasks. This is clearly demonstrated in Figure~\ref{fig:v-star}, which shows that applying image partitioning on V* Bench boosts the average accuracy to \textbf{68.5\%} from the \textbf{49.6\%} achieved when processing the image directly. The figure also illustrates a consistent performance improvement as the number of sub-images increases, confirming that this approach enhances the model's capacity for detailed visual analysis and effectively mitigates the performance degradation caused by token reduction on detail-oriented tasks

% \subsection{Impact and Mitigation of Token Compression}
% \label{sec:token}

% To assess token compression's impact on image understanding and explore mitigation strategies for tasks sensitive to token reduction, we conduct experiments on five general benchmarks and V* Bench~\citep{vstar}, which specifically evaluates a model's ability to localize small objects within large images.

% \paragraph{Impact of Token Compression} As shown in Figure~\ref{fig:token_number}, setting the token count to 144 per image significantly reduces inference cost while maintaining overall performance, with the effect being especially notable on SEEDBench.

% \paragraph{Mitigation Strategies} Figure~\ref{fig:v-star} demonstrates that increasing the number of sub-images initially brings substantial performance improvements, indicating enhanced fine-grained image understanding. Furthermore, as further evidenced in the table, partitioning images into sub-images effectively mitigates the performance drop caused by token compression on fine-grained tasks. Notably, average accuracy rises markedly from 49.6 to 68.5 when using image partitioning rather than processing the entire image directly.

% \vspace{-1mm}
\begin{table}[t]
% \vspace{-2mm}
    \centering
    \footnotesize
    \setlength{\tabcolsep}{4pt}
    \begin{tabular}{l|c|cc}
        \toprule
        \textbf{Model}&  \textbf{Size}& \textbf{VQA-RAD} &\textbf{PathVQA} \\
        \midrule
        GPT-4V  &  - & 39.5 &- \\
        LLaVA &  34B & 58.6 &59.1\\
        LLaVA-Med &  7B & 55.5 &35.9 \\
        HuatuoGPT-V &  8B & 63.8 &\textbf{59.9} \\
        \midrule
        LongLLaVA-Med & 9B&\textbf{ 68.5} &55.0\\
        \bottomrule
    \end{tabular}
    % \vspace{-1mm}
    \caption{Comparison of model performance on pathology image understanding benchmarks.}
    % \vspace{-2mm}
    \label{tab:app1-1}
\end{table}

\begin{table}[t]
    \centering
    \footnotesize
    \setlength{\tabcolsep}{4pt}
    \begin{tabular}{l|cccc}
        \toprule
        \textbf{Model} & \textbf{Acc.} & \textbf{Rec.} & \textbf{Prec.} & \textbf{F1}\\
        \midrule
        CT-CLIP  & 65.1 & 73.8 & 30.4 & 43.0 \\
        LongLLaVA-Med & \textbf{86.7} & \textbf{77.6} & \textbf{35.5} & \textbf{48.5}\\
        \bottomrule
    \end{tabular}
    % \vspace{-1mm}
    \caption{Model performance on the 3D CT image interpretation task. Acc., Rec., and Prec. denote Accuracy, Recall, and Precision, respectively.}
    \label{tab:app1-2}
    % \vspace{-4mm}
\end{table}

\section{Applications}
\subsection{Applications in Healthcare}

To showcase LongLLaVA’s effectiveness in healthcare, we introduce \textbf{LongLLaVA-Med}, a model derived by fine-tuning LongLLaVA-9B on the PubMedVision dataset~\citep{chen2024huatuogpt}. This is a large-scale dataset comprising 1.3 million medical VQA samples, which was constructed by employing an "unblinded" Multimodal Large Language Model (MLLM) to denoise and reformat raw image-text pairs from biomedical literature. We evaluate the resulting model's capabilities in two critical tasks: pathology image analysis and 3D CT image interpretation.

\paragraph{Pathology Image Understanding.} Pathology image analysis demands both fine-grained visual recognition and a deep understanding of medical knowledge. We evaluate LongLLaVA-Med on two benchmarks: VQA-RAD~\citep{lau2018dataset} and PathVQA~\citep{he2020pathvqa}. As shown in Table~\ref{tab:app1-1}, our model achieves competitive performance compared to state-of-the-art approaches, despite being trained on less data.

\paragraph{3D CT Image Interpretation.}
To test its 3D vision capabilities, we apply LongLLaVA-Med to CT scan interpretation. Each 3D CT scan, consisting of multiple slices, is processed as a sequence of RGB images. We conduct zero-shot evaluation on the CT-RATE~\citep{hamamci2024foundation} validation set, which includes 1,304 samples with varying resolutions ($512 \times 512$ to $1024 \times 1024$, average 690) and slice counts (100–984, average 300). As shown in Table~\ref{tab:app1-2}, LongLLaVA-Med surpasses previous state-of-the-art results by 21.6\%, setting a new benchmark for 3D CT image interpretation.

% \begin{table}[t]
%     \footnotesize
%     \centering
%     % \vspace{-1mm}
%     \addtolength\tabcolsep{-8pt}
%     \begin{tabular}{lccc}
%         \toprule
%         \textbf{Model} & \textbf{LLaVA1.5-7B} & \textbf{GeoChat-7B} & \textbf{LongLLaVA-9B} \\
%         \midrule
%         Zero-shot & 58.6 & 53.5 & \textbf{65.2} \\
%         \midrule
%         \textbf{Model} & \textbf{SkySenseGPT-7B}  &  & \textbf{LongLLaVA-RS*-9B}  \\ \midrule
%         Fine-tuned & 79.8  &  & \textbf{82.3}  \\
%         % \midrule
%         % \textbf{Model} & \textbf{SkySenseGPT} & \textbf{LongLLaVA-RS*} &  \\
%         \bottomrule
%     \end{tabular}
%     % \vspace{-2mm}
%     \caption{Results on FIT-RSFG-VQA}
%     % \vspace{-4mm}
%     \label{tab:app2-1}
% \end{table}

\begin{figure}[t]
% \vspace{-2mm}
    \centering
    % \vspace{-3mm}
    %\vspace{-9.5pt}
    \includegraphics[width=1.0\linewidth]{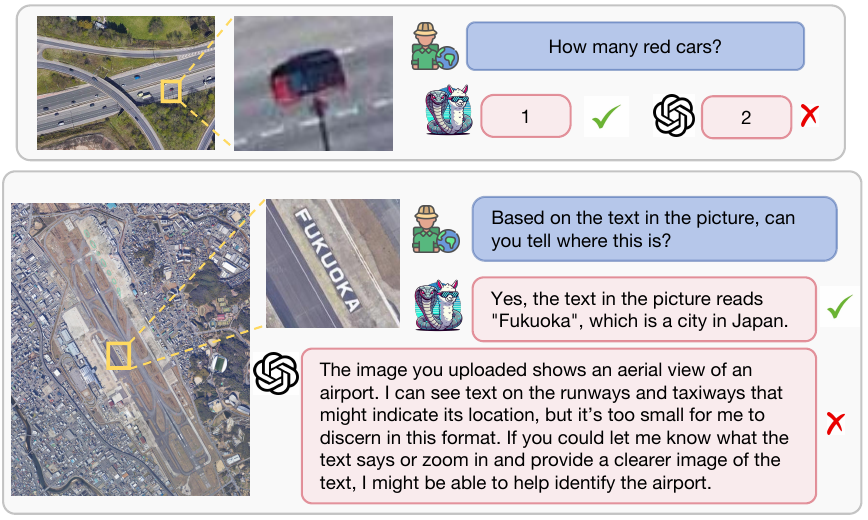}
    % \vspace{-7mm}
    \caption{Comparative Study of Remote Sensing on the STAR Dataset.}
    % \vspace{-1mm}
    \label{fig:app2-1}
\end{figure}

\begin{table}[t]
    \centering
    \small
    \begin{tabular}{lc}
        \toprule
        \textbf{Model} & \textbf{Score (\%)} \\
        \midrule
        \rowcolor{gray!15} \multicolumn{2}{c}{\textbf{Zero-shot Evaluation}} \\
        \quad LLaVA-1.5 (7B) & 58.6 \\
        \quad GeoChat (7B) & 53.5 \\
        \quad LongLLaVA (9B) & \textbf{65.2} \\
        \midrule
        \rowcolor{gray!15} \multicolumn{2}{c}{\textbf{Fine-tuned Evaluation}} \\
        \quad SkySenseGPT (7B) & 79.8 \\
        \quad LongLLaVA-RS* (9B) & \textbf{82.3} \\
        \bottomrule
    \end{tabular}
    \caption{Results on FIT-RSFG-VQA. The best performance in each category is highlighted in \textbf{bold}.}
    \label{tab:app2-1}
\end{table}

\subsection{Applications in Science}

In the scientific domain, we focus on geology and the interpretation of remote sensing imagery, which requires models to perform VQA on high-resolution satellite images~\citep{zhou2024towards}. Following the recent work of SkySenseGPT~\citep{luo2024skysensegpt}, a state-of-the-art MLLM for this field, we adopt the FIT-RSFG-VQA task~\citep{luo2024skysensegpt} to evaluate models on fine-grained perception and instruction-following abilities.

As shown in Table~\ref{tab:app2-1}, LongLLaVA exhibits strong performance among all evaluated models. Notably, after fine-tuning on only 27\% of the SkySenseGPT data, LongLLaVA surpasses existing state-of-the-art models.

To address the resolution limitations of FIT-RSFG-VQA ($512 \times 512$ pixels), we further evaluate on two high-resolution images from the STAR dataset~\citep{li2024star}, with resolutions of $1024 \times 768$ and $3327 \times 4083$. This enables a more comprehensive assessment of model capabilities. As illustrated in Figure~\ref{fig:app2-1}, LongLLaVA effectively answers fine-grained VQA queries by segmenting large images into manageable subimages, consistently outperforming GPT-4V, especially on tasks requiring detailed visual analysis.

% To highlight LongLLaVA's high-resolution remote sensing capabilities beyond FIT-RSFG-VQA's 512×512 resolution, we tested two STAR images~\citep{li2024star} (1024×768 and 3327×4083). As shown in Figure~\ref{fig:app2-1}, 

% \vspace{15mm}

% \subsection{Multi-modal agent}

% MotionLLM
% Web agent/ Mobile agent; or synthesized data;

% long 

% % Adaptive patching for 

% % high-resolution to muitlple image  problem
% scaling law?

% \vspace{2mm}
\section{Conclusion}
% \vspace{-2mm}

In this study, we introduce LongLLaVA, an innovative hybrid architecture model that excels in long-context multi-modal understanding. The model integrates Mamba and Transformer blocks, leveraging temporal and spatial dependencies between multiple images to construct data, and employs a progressive training strategy. LongLLaVA demonstrates competitive performance across various benchmarks while ensuring efficiency, setting a new standard for long-context MLLMs.

\section*{Limitations}
While our current model achieves a multimodal context length of 176K tokens, this is still limited compared to the ideal context range of 10–100 million tokens, which would enable more comprehensive understanding of large-scale inputs. Extending the context window to this scale remains a significant technical challenge, involving issues such as computational efficiency and memory constraints. Further research is needed to explore more effective architectures and optimization strategies to address these limitations.

\section*{Acknowledgments}
% \vspace{-2mm}

This work was supported by Major Frontier Exploration Program (Grant No. C10120250085) from the Shenzhen Medical Academy of Research and Translation (SMART), the Shenzhen Science and Technology Program (JCYJ20220818103001002), NSFC grant 72495131, Shenzhen Doctoral Startup Funding (RCBS20221008093330065), Tianyuan Fund for Mathematics of National Natural Science Foundation of China (NSFC) (12326608), Shenzhen Science and Technology Program (Shenzhen Key Laboratory Grant No. ZDSYS20230626091302006), and Shenzhen Stability Science Program 2023.

% \section*{Accessibility}
% Authors are kindly asked to make their submissions as accessible as possible for everyone including people with disabilities and sensory or neurological differences.
% Tips of how to achieve this and what to pay attention to will be provided on the conference website \url{http://icml.cc/}.

% \section*{Software and Data}

% If a paper is accepted, we strongly encourage the publication of software and data with the
% camera-ready version of the paper whenever appropriate. This can be
% done by including a URL in the camera-ready copy. However, \textbf{do not}
% include URLs that reveal your institution or identity in your
% submission for review. Instead, provide an anonymous URL or upload
% the material as ``Supplementary Material'' into the OpenReview reviewing
% system. Note that reviewers are not required to look at this material
% when writing their review.

% % Acknowledgements should only appear in the accepted version.
% \section*{Acknowledgements}

% \section*{Impact Statement}

% \newpage
\clearpage
\newpage
\bibliography{custom}

%%%%%%%%%%%%%%%%%%%%%%%%%%%%%%%%%%%%%%%%%%%%%%%%%%%%%%%%%%%%%%%%%%%%%%%%%%%%%%%
%%%%%%%%%%%%%%%%%%%%%%%%%%%%%%%%%%%%%%%%%%%%%%%%%%%%%%%%%%%%%%%%%%%%%%%%%%%%%%%
% APPENDIX
%%%%%%%%%%%%%%%%%%%%%%%%%%%%%%%%%%%%%%%%%%%%%%%%%%%%%%%%%%%%%%%%%%%%%%%%%%%%%%%
%%%%%%%%%%%%%%%%%%%%%%%%%%%%%%%%%%%%%%%%%%%%%%%%%%%%%%%%%%%%%%%%%%%%%%%%%%%%%%%
\newpage

\appendix

% \section*{Appendix Table of Contents}
% \startcontents[sections]
% \printcontents[sections]{l}{1}{\setcounter{tocdepth}{2}}

\newpage

\section{Details about Hybrid Motivation}
\label{sec:hybrid}

This section covers the details about investigating the respective strengths and limitations of various model architectures concerning in-context learning (ICL) capabilities and inference efficiency. Our analysis aims to underscore the advantages offered by hybrid architectures, which are designed to amalgamate the beneficial characteristics of their constituent architectural paradigms.

\noindent\textbf{Baselines} To assess the Mamba architecture, we employed the Falcon-mamba model~\citep{zuo2024falcon} featuring 7.3 billion parameters. This model was trained and evaluated under identical settings to our proposed model. Despite the inherent challenges in precisely aligning parameter counts across different Multimodal Large Language Models (MLLMs), Falcon-mamba represents the largest publicly available Mamba configuration suitable for this comparative analysis. For the Transformer architecture, we selected LLaVA-1.6~\citep{liu2024llavanext} with 13B parameters. This choice was motivated by its inference parameter consistency with LongLLaVA, thereby facilitating a more equitable comparison of inference efficiency.

\noindent\textbf{ICL Analysis Settings} We conducted an evaluation of multimodal in-context learning performance using the ``Matching Image'' task from the VL-ICL benchmark~\citep{zong2024vliclbenchdevildetails}. In this task, the input consists of an image pair, denoted as $x = \{x_1, x_2\}$. The corresponding output, $y$, signifies whether a predefined relation, $r$, is present between the two images. The objective for the MLLMs is to discern this underlying relation from a set of provided examples.

\noindent\textbf{Efficiency Analysis Settings} Our examination of inference efficiency encompasses four key metrics: Prefill Time (latency of the initial inference step), Throughput (measured as subsequent tokens generated per second), Memory Usage, and \revise{Maximum Throughput (defined as the throughput achieved under the maximum possible batch size)}. For these evaluations, we standardized the input text length to 100,000 tokens. We then measured the time taken and the peak memory consumption for generating outputs of 1 token and 1,000 tokens. Throughput was subsequently calculated using the formula: $(1000-1) / (\text{time}_{1000} - \text{time}_{1})$. To more accurately reflect real-world application scenarios, both Transformer and Hybrid architectures were benchmarked utilizing the vLLM framework~\citep{kwon2023efficientmemorymanagementlarge} and \texttt{Int8} quantization~\citep{frantar2023gptqaccurateposttrainingquantization}.

\begin{table*}[t]\footnotesize
    \centering
    \begin{tabularx}{\textwidth}{l *{6}{>{\centering\arraybackslash}X}}
        \toprule
        \textbf{Downcycling Strategy} & \textbf{Arithmetic Mean} & \textbf{Spherical Mean} & \textbf{Expert-0} & \textbf{Expert-5} & \textbf{Expert-12} & \textbf{Expert-15} \\
        \midrule
        MMLU & 52.7 & \textbf{53.2} & \textbf{53.2} & 51.9 & 52.6 & 52.2 \\
        Aft. Train & 53.8 & \textbf{54.3} & \textbf{54.3} & 53.3 & 53.8 & 53.3 \\
        \midrule
        BBH & 36.7 & 36.7 & 37.2 & 36.7 & \textbf{37.4} & 36.3 \\
        Aft. Train & 37.8 & 37.9 & 38.4 & \textbf{38.9} & \textbf{38.9} & 37.9 \\
        \bottomrule
    \end{tabularx}    
    \caption{Performance of Different Downcycling Strategies on MMLU and BBH}
    \vspace{-4mm}
    \label{tab:expert-select}
\end{table*}

\section{Experiments Settings for Hybrid Ratio}
\label{sec:ratio}

To investigate the optimal ratio of Attention to Mamba layers (denoted as $a:m$) within 1.3B parameter models, various model configurations were trained using 250B tokens randomly sampled from the FineWeb-Edu dataset~\citep{penedo2024finewebdatasetsdecantingweb}. The training utilized a global batch size of 512, a cosine learning rate schedule with a peak learning rate of $1 \times 10^{-4}$, and the AdamW optimizer (with $\beta_1=0.9$, $\beta_2=0.95$, and a weight decay of 0.1). Upon completion of training, model performance was evaluated on multiple benchmark datasets.

The evaluation benchmarks included:
\begin{itemize}
    \item \textbf{HellaSwag}~\citep{zellers2019hellaswagmachinereallyfinish}: Evaluated using a 10-shot setting, this benchmark assesses commonsense inference by requiring models to choose the most plausible continuation of a given context from four options.
    \item \textbf{ARC-Challenge}~\citep{clark2018thinksolvedquestionanswering}: This benchmark was evaluated with a 25-shot methodology and tests complex question-answering abilities, particularly the more difficult questions from the AI2 Reasoning Challenge (ARC) which often demand multi-step reasoning.
    \item \textbf{Natural Questions (NQ)}~\citep{kwiatkowski-etal-2019-natural}: A 5-shot evaluation was used for this benchmark, which measures the model's capacity to answer real user questions from Google Search without access to external documents, relying on its internal knowledge.
    \item \textbf{BoolQ}~\citep{clark2019boolqexploringsurprisingdifficulty}: Evaluated in a 10-shot setup, this benchmark assesses reading comprehension through yes/no questions paired with short passages, where the model must determine the answer's veracity based on the text.
\end{itemize}

\section{Preliminary Experiments on Expert Selection for LongLLaVA-9B}
\label{app:expert}
% \vspace{-2mm}

\revise{To determine the optimal expert selection method in the MoE layers we also conducted preliminary experiments. Using prevalent LLM benchmarks, MMLU~\citep{hendrycks2020measuring} and BBH~\citep{suzgun2022challenging}, we evaluated three expert selection strategies: numerical averaging, spherical averaging, and random expert selection.}

\revise{These methods were compared both before and after Pure-text Instruction Tuning with dataset of 278k pure-text entries, aggregated from Evol-instruct-GPT4~\citep{xu2023wizardlmempoweringlargelanguage}, WildChat~\citep{zhao2024wildchat1mchatgptinteraction}, alongside LongAlign~\citep{bai2024longalignrecipelongcontext}. As shown in Table~\ref{tab:expert-select}, the differences in model performance were minimal across the selection methods. Therefore, for simplicity, we opted to use Expert-0.}

\section{Details of Multi-Image Evaluation}
\label{app:eval}

\subsection{Benchmarks}
The multimodal long-context understanding capabilities of our model are primarily assessed using five multi-image benchmarks. These include MileBench~\citep{song2024milebenchbenchmarkingmllmslong}, selected for its focus on multimodal long-context scenarios. For video analysis, we additionally incorporate Video-MME~\citep{fu2024video}, MVBench~\citep{li2024mvbenchcomprehensivemultimodalvideo}, and LongVideoBench~\citep{wu2024longvideobenchbenchmarklongcontextinterleaved}. Detailed descriptions of these benchmarks are provided subsequently. 

\paragraph{Multi-image Benchmarks}
To evaluate multi-image understanding capabilities, the following benchmarks were employed:
\begin{itemize}
    \item \textbf{MileBench}~\citep{song2024milebenchbenchmarkingmllmslong}: This benchmark evaluates performance in long-context scenarios, with a particular emphasis on its Temporal, Semantic, and Information Retrieval (IR) components.
    \item \textbf{Video-MME}~\citep{fu2024video}: This benchmark assesses video analysis capabilities across 30 distinct sub-fields. The evaluation protocol typically involves processing 128 frames uniformly sampled from each video, without relying on subtitle information.
    \item \textbf{MVBench}~\citep{li2024mvbenchcomprehensivemultimodalvideo}: MVBench targets 20 challenging video understanding tasks that are intractable with single-frame analysis, thus requiring multi-frame reasoning.
    \item \textbf{LongVideoBench}~\citep{wu2024longvideobenchbenchmarklongcontextinterleaved}: This benchmark provides a question-answering (QA) framework with interleaved video-language inputs, where video durations can extend up to one hour.
\end{itemize}

\subsection{Comparative Models}

Our model is benchmarked against a comprehensive suite of existing models, encompassing three commercial and thirteen open-source counterparts. The commercial models include GPT-4V\footnote{\texttt{gpt-4-vision-preview}}~\citep{openai2024gpt4technicalreport}, GPT-4o\footnote{\texttt{gpt-4o-2024-08-06}}, Claude3-Opus~\citep{TheC3} and Gemini-1.5-Pro\footnote{\texttt{gemini-1.5-pro}}~\citep{geminiteam2024gemini15unlockingmultimodal}. The open-source models comprise Qwen2-VL2~\citep{wang2024qwen2vlenhancingvisionlanguagemodels}, Qwen2.5-VL~\citep{bai2025qwen25vltechnicalreport}, InternVL2~\citep{chen2024far}, InternVL2.5~\citep{chen2024expanding}, Phi-3-Vision~\citep{abdin2024phi3technicalreporthighly}, OmChat~\citep{zhao2024omchat}, \revise{LongVA}, LongVILA~\citep{xue2024longvilascalinglongcontextvisual}, Video-LLaMA-2~\citep{cheng2024videollama2advancingspatialtemporal}, Cobra~\citep{zhao2025cobraextendingmambamultimodal}, Mini-Gemini~\citep{li2024minigeminiminingpotentialmultimodality}, mPLUG-Owl3~\citep{ye2024mplugowl3longimagesequenceunderstanding}, and VideoChat2~\citep{li2024mvbenchcomprehensivemultimodalvideo}. For consistent and reproducible evaluations, the temperature parameter is set to \texttt{0}.

\section{Details of Single-Image Evaluation}
\label{app:single_eval}

The single-image evaluation is designed to investigate the model's fundamental capabilities and the impact of extended long-context training on single-image understanding.

\subsection{Experimental Setup}
We employed a comprehensive suite of benchmarks to assess various aspects of visual understanding and cognitive processing within a single-image context. These benchmarks include GQA~\citep{hudson2019gqa}, MME~\citep{fu2023mme}, MM-Vet~\citep{yu2023mmvet}, ScienceQA~\citep{lu2022learn}, SEED-Bench-v1~\citep{li2023seedbench}, MMBench~\citep{liu2023mmbench}, MMMU~\citep{yue2024mmmumassivemultidisciplinemultimodal}, BLINK~\citep{fu2024blinkmultimodallargelanguage}, \revise{ChartQA~\citep{masry2022chartqabenchmarkquestionanswering}, and DocVQA~\citep{mathew2021docvqadatasetvqadocument}}. Detailed descriptions are provided below.

\begin{table*}[h]\footnotesize
% %%\vspace{-4mm}
\centering
\addtolength\tabcolsep{-4.7pt} 
\begin{tabular}{c|rrc|cccccccccc}
\toprule
Model & TFLOPs & \revise{\#P} & \#T & \revise{ChartQA} & \revise{DocVQA}& GQA & MM-Vet & MME$^P$  & MMB & MMMU & SQA$^{I}$ & SEED$^{v1}_{img}$ & BLINK\\ \midrule
\rowcolor{gray!15} \multicolumn{14}{c}{\textbf{Proprietary Models}} \\
GPT-4V  & -& - & -& 75.6& - & - & 67.7 & 1926.5 & \textbf{81.3} & \textbf{56.8} & \textbf{82.1} & \textbf{69.1} & \\   
Gemini-1.5  & - & -& - & \textbf{81.3}& \textbf{90.9}& - & 65.8 & \textbf{2148.9} & 73.6 & 48.9 & 81.4 & 62.9 & \\  
Claude3-Opus  & - & -& -  & 80.8& 89.3& - & \textbf{74.2} & 1586.8 & 63.3 & 54.9 & - & 42.0 & \\ 
\midrule
\rowcolor{gray!15}\multicolumn{14}{c}{\textbf{Open-source MLLMs}} \\ 
InternVL2 & 5.45 & 8B & 576 & 83.3 & 91.6  & - & - & 2210.3	 & 82.9 & 52.6 & - & - & 50.9\\
InternVL2.5 & 5.45 & 8B & 576 & 84.8 & 93.0 & - & - & 2344.1 & 83.8 & 56.0 & - & - & 54.8 \\
OmChat & 5.18 & 8B & 576 & -& -  & - & 39.6 & - & 78.8 & 45.9 & - & - & -\\
LongVILA & 5.18 & 8B & 576 & -& -  & 65.4 & 51.7 & - & 83.4 & - & - & 70.6 & -\\
Qwen2-VL & 5.05 & 7B & 576 & 83.0 & 94.5  & - & - & 1872.0 &  & 54.1 & - & - &\\
Qwen2.5-VL & - & 7B & - & 87.3 & 95.7 & - & - & 2347.0 & 83.5 & 58.6 & - & - & 56.4 \\
\midrule
\rowcolor{gray!15}\multicolumn{14}{c}{\textbf{Open-source Efficient MLLMs}} \\ 
Phi-3-Vision & 3.56 & 4B &576 & 81.8 & 69.3  & - & - & - & 80.5 & 40.4 & 90.8 & - & - \\
Cobra & 2.35 & 7B & 768 & - & -  & 63.9 & - & 1496.5 & - & 37.2 & - & - &\\
LongLLaVA-9B  & 0.58 & 9B &144 & 44.8 & 47.4  & 58.4 & 32.3 & 1504.6 & 65.6 & 34.4 & 69.9 & 67.9  & 50.2 \\
LongLLaVA-A13B  & 0.86 & 53B & 144 & 46.3 & 51.2  & 59.9
& 35.2 & 1523.9 & 63.7 & 39.2 & 73.4 & 65.3 & 52.4\\
\midrule
\textbf{LongLLaVA-9B*}  & 4.86 & 9B & - & 72.3 & 83.6  & 72.3 & 42.6 & 1693.6 & 72.8 & 45.3 & 76.3 & 70.9  & 52.2 \\
\textbf{LongLLaVA-A13B*}  & 5.14 & 53B & - & 81.6 & 90.8  & 76.5 & 53.6  & 1823.9 & 79.4 & 52.5 & 80.4 & 72.4 & 55.2 \\
% Jamba-int8-2D-A13B & LongLLaVA-LenPack & 144 & -
% & - & - & - & - & - & - \\
% Jamba-int8-2D-A13B & LongLLaVA-LenPack & Fit & -
% & - & - & - & - & - & - \\
\bottomrule
\end{tabular}
\caption{Single-image Evaluation. TFLOPs represents the number of floating-point operations required to infer 1 images. The highest scores for proprietary and open-source MLLMs are marked in bold. \#Token refers to the token count for one image. * means using Mitigating Token Compression Strategy mentioned in Section~\ref{sec:token}.} 

\label{tab:single_image} 
\end{table*}

\begin{table*}[h]
    \centering
    % \vspace{-4mm}
    \small % or \scriptsize for even smaller text
    \begin{tabularx}{\textwidth}{l*{7}{>{\centering\arraybackslash}X}}
    \toprule
    & \textbf{MMLU} & \textbf{BBH} & \textbf{GQA} & \textbf{MMMU} & \textbf{SQA$^{I}$} & \textbf{SEED$^{v1}_{img}$} & \textbf{Mile$^{*}_{avg}$} \\
    \midrule
    \textbf{LongLLaVA-9B} & \textbf{53.9} & \textbf{38.8} & \textbf{58.4} & \textbf{34.4} & \textbf{69.9} & \textbf{67.9} & 46.5 \\
    \textbf{w/o Replay Data} & 52.3 & 36.2 & 57.5 & 31.2 & 53.5 & 64.3 & 46.8 \\
    \textbf{Replace with Multi-Image} & 52.6 & 35.9 & 57.2 & 29.8 & 52.6 & 63.8 & \textbf{47.2} \\
    \bottomrule
    \end{tabularx}
    \caption{Comparison of Model Performance With and Without Replay Data.}
    % \vspace{-2mm}
    \label{tab:replay-comparison}
\end{table*}

\paragraph{Single-Image Benchmarks}
To evaluate the model's single-image understanding capabilities, we selected eight commonly utilized benchmarks. These are:
\begin{itemize}
    \item \textbf{GQA}~\citep{hudson2019gqa}: A benchmark for real-world visual reasoning and compositional question answering.
    \item \textbf{MME}~\citep{fu2023mme}: A comprehensive benchmark for evaluating multimodal perception and cognition; the perception-focused subset was employed in this study.
    \item \textbf{MM-Vet}~\citep{yu2023mmvet}: Examines six core visual-linguistic (VL) capabilities alongside sixteen integrated tasks derived from these capabilities.
    \item \textbf{ScienceQA}~\citep{lu2022learn}: Comprises 4,210 questions on diverse science topics, featuring detailed annotations.
    \item \textbf{SEED-Bench-v1}~\citep{li2023seedbench}: Evaluates multimodal comprehension across twelve dimensions in both image and video modalities; our analysis utilized the image-based subset.
    \item \textbf{MMBench}~\citep{liu2023mmbench}: A systematically designed benchmark covering twenty distinct multimodal ability dimensions.
    \item \textbf{MMMU}~\citep{yue2024mmmumassivemultidisciplinemultimodal}: Assesses multimodal models on multidisciplinary tasks requiring university-level expertise, spanning 183 subfields and 30 types of images.
    \item \textbf{BLINK}~\citep{fu2024blinkmultimodallargelanguage}: A benchmark for multimodal LLMs that specifically targets core visual perception abilities not emphasized in existing evaluations.
\end{itemize}

\paragraph{Comparison Models}
Our model was benchmarked against a comprehensive suite of existing models, comprising four commercial and thirteen open-source alternatives. This set of comparison models is identical to that used in the Multi-Image evaluations. For consistent and reproducible evaluations, the temperature parameter is set to \texttt{0}.

\subsection{Results Analysis}
As shown in Table~\ref{tab:single_image}, for the single-image understanding task, the LongLLaVA series models, when using default inference settings, achieve performance comparable to other efficient multimodal models, while requiring fewer inference FLOPs. However, their performance still lags behind that of some leading multimodal models, primarily due to token compression.

To fully realize the potential of LongLLaVA in single-image understanding, we applied the token compression mitigation strategy described in Section~\ref{sec:token}. Specifically, we pad each image so that its height and width are multiples of 168, then partition it into sub-images of size $168 \times 168$. This approach effectively eliminates the adverse effects of token compression. Experimental results demonstrate that, with this mitigation strategy, the LongLLaVA series achieves performance on par with state-of-the-art multimodal models.

\begin{table}[t]
    \centering
    \small % Use \scriptsize for even smaller text if necessary
    \begin{tabular}{lcc}
    \toprule
    & \textbf{MMLU} & \textbf{BBH} \\
    \midrule
    \textbf{LongLLaVA-9B (w/o Replay Data)} & 52.3 & 36.2 \\
    \textbf{with 10K} & 52.9 & 37.3 \\
    \textbf{with 20K} & 53.4 & 38.1 \\
    \textbf{with 50K} & \textbf{53.9} & 38.8 \\
    \textbf{with 100K} & 53.9 & \textbf{39.2} \\
    \bottomrule
    \end{tabular}
    % \vspace{-4mm}
    \caption{Impact of Text Replay Data Quantity.}
    \label{tab:text-replay}
\end{table}

% \vspace{-2mm}
\section{Replay Data Ablation Study}
\label{app:abl-2}

% \vspace{-2mm}
To assess the impact of replay data, we conducted three experiments as part of the Replay Data Ablation Study.

\paragraph{Comparison With and Without Replay Data.}
\revise{We first conducted experiments comparing models trained with and without replay data. To isolate the effect of replay data from the impact of increased training data, we performed an ablation study by replacing replay data in the original training recipe with an equivalent amount of multi-image data. The results, presented in Table~\ref{tab:replay-comparison}, demonstrate that \textbf{replay data is essential for preserving the model's original single-image understanding and text-following capabilities}.}

\paragraph{Replay Data Quantity Ablation.}

\begin{table}[t]\small
    \centering
    \addtolength\tabcolsep{-5.0pt}
    \small % Use \scriptsize for even smaller text if necessary
    \begin{tabular}{lccccc}
    \toprule
    & \textbf{GQA} & \textbf{MMMU} & \textbf{SQA$^{I}$} & \textbf{SEED$^{v1}_{img}$} & \textbf{Mile$^{*}_{avg}$} \\
    \midrule
    \textbf{w/o Replay Data} & 57.5 & 31.2 & 53.5 & 64.3 & \textbf{46.8} \\
    \textbf{with 50K} & 57.9 & 32.3 & 58.2 & 66.2 & 46.5 \\
    \textbf{with 100K} & 57.9 & 33.5 & 62.7 & 67.1 & 46.5 \\
    \textbf{with 200K} & 58.2 & 34.5 & 67.1 & 67.9 & \textbf{46.8} \\
    \textbf{with 400K} & \textbf{58.5} & \textbf{35.2} & \textbf{73.2} & \textbf{68.2} & 46.4 \\
    \bottomrule
    \end{tabular}
    \caption{Impact of Single-Image Replay Data Quantity.}
    \label{tab:single-image-replay}
\end{table}

\revise{We also examined the impact of varying the quantity of replay data.
For \textbf{text replay data}, the supplementary experiments reveal that adding text replay data enhances the model's text-following ability, although the improvement eventually saturates, as shown in Table~\ref{tab:text-replay}. For \textbf{single-image replay data}, the results in Table~\ref{tab:single-image-replay} indicate that the model's single-image capability continues to improve with increased data volume and has not yet reached saturation. However, the improvement in multi-image tasks is limited.}

% \section{\revise{Additional Application}}

\end{document}